# Ancient Korean Archive Translation:
# Comparison Analysis on
# Statistical phrase alignment, LLM in-context learning,
# and inter-methodological approach

Sojung Lucia Kim
Seoul National University
Nara.Labs
sojung.kim@snu.ac.kr
sojung.kim@narainformation.com

Taehong Jang
Nara.Labs
starbirdnara@narainformation.com

Joonmo Ahn
Nara.Labs
jmahn.nara@narainformation.com

## Abstract

This study aims to compare three methods for translating ancient texts with sparse corpora: (1) the traditional statistical translation method of phrase alignment, (2) in-context LLM learning, and (3) proposed inter methodological approach - statistical machine translation method using sentence piece tokens derived from unified set of source-target corpus. The performance of the proposed approach in this study is 36.71 in BLEU score, surpassing the scores of SOLAR-10.7B context learning and the best existing Seq2Seq model. Further analysis and discussion are presented.

## 1 Introduction

This study aims to compare three methods for translating ancient texts with sparse corpora: (1) the traditional statistical translation method of phrase alignment, (2) in-context learning using large language models (LLMs), and (3) proposed inter methodological approach - statistical translation method using Sentence Piece tokens derived from unified set of source-target corpus. The study implements these methods and presents a comparative analysis of the results.

## 2 Corpus

### 2.1 The Annals of the Joseon Dynasty (Joseon Wangjo Sillok)

The parallel corpus used in this study consists of the original texts of the Annals of the Joseon Dynasty in classical Chinese and their translated versions in modern Korean.

The Annals of the Joseon Dynasty are a comprehensive chronicle of the history of the Joseon Dynasty (1392-1910). They detail various aspects of politics, society, economy, and culture during the reigns of the 25 kings of the Joseon Dynasty. Consisting of a total of 1,893 volumes, the Annals are considered one of the most extensive historical records in the world. The source data in this corpus were written in Classical Chinese, characterized by several unique features. The characters used include those that are rarely used in modern contexts or are specific to Korean usage. Additionally, the text lacks spacing between words, uses special conventions such as using "上" to refer to the king, and leaves a blank space before the names of important individual to show respect toward those individuals. It consists of daily records written in a diary format. They are primarily composed of conversations between the king and his officials.

The Annals of the Joseon Dynasty were translated over approximately 20 years by the Foundation for the Translation of Korean Classics and are made publicly available through the National Institute of Korean History[1].

### 2.2 Data Preparation

We collected parallel corpus through web scraping from the National Institute of Korean History for three months. The entire dataset comprises approximately 64 million characters and each of records ages from short passages of one or two words to longer paragraphs of around several thousand characters. Since this classical parallel corpus involves translating Chinese characters into Korean, Korean tends to be much longer compared to classical Chinese. Therefore, we restricted our selection to diaries consisting of up to 1024 characters in Korean and 128 characters in Chinese, which yield to 252,773 records.

---

[1] https://sillok.history.go.kr/

## 3 Machine Translation

| Type | Number of Records |
|---|---|
| Train | 202,218 |
| Test | 50,555 |
| Total | 252,773 |

Table 1: Corpus Size

In this section, we explain three machine learning approaches that can be used for classical translation: statistical phrase alignment, Seq2Seq, and LLM in context learning and introduce the performance scores known up to the present.

### 3.1 Statistical phrase alignment

The goal of machine translation is to find the translation, $\hat{t}$ which is defined as:

$$\hat{t} = \arg max_t \, p(t|s)$$

where $p(t|s)$ is the probability model. The argmax implies a search for the best translation $\hat{t}$ in the space of possible translations t.

This method relies on well-aligned corpora. It tokenizes both the source and target sentences, scoring each token based on how well it matches. Then, it constructs language models, such as 3-gram models, based on the target sentences. When given a source sentence, it generates translation candidates composed of the best-matching tokens. These candidates are then evaluated using the trained language model to select the most fluent sentence (Hoang and Kohen, 2008). Using the Moses toolkit on the WMT corpus, the authors achieved a BLEU score of 19.57 for French-English translation on a dataset of 2000 sentences.

### 3.2 Sequence to Sequence Model

This method involves generating sequences from input sequences that belong to a different domain. Translation is a prominent example of this model. There are various studies on classical translation, and one notable example is the research conducted by Park et al. (2020). The authors presented an ensemble model, achieving the highest score of BLEU 32.57.

### 3.3 LLM in - context learning

The rich adaptability of LLMs can be attributed to their template-based emergent capabilities, as described by Brown et al. (2020).

Specifically, the prompt is made up of in-context exemplars $(X_i, Y_i)_{i=1}^{k}$ and in-context template $T$. Exemplars are often picked from supervised data, where $Y_i$ is the ground truth corresponding to the input sentence $X_i$. Template $T$ is usually a human-written instruction related to the target task. Wrapping exemplars with the template and concatenating them together produce the final prompt:

$$P = \text{T}(X_1, Y_1) \oplus \text{T}(X_2, Y_2) \oplus \cdots \\ \oplus \text{T}(X_k, Y_k)$$

where $\oplus$ denotes the concatenation symbol, e.g., whitespace, line-break. During inference, LLM is able to generate the corresponding output Y of the test sample X under the guidance of the prompt:

$$\arg \max_y p(P \oplus \text{T}(X, Y))$$

Recent research by Zhu et al. (2023) utilized LLMs for translation, applying various forms of prompts to multiple LLMs and comparing their translation performance. In those authors work, a staggering 102 languages and 606 translation directions were tested. For instance, when translating from the Germanic languages to English, the highest BLEU score of 48.51 was achieved using GPT-4 as the reference model.

## 4 Proposed Method – Inter Methodological Approach

Despite the success of machine learning and LLMs, translation performance remains low (Zhu et al., 2023), particularly in cases like classical translation where the corpus volume is inherently limited. Moreover, it's crucial to recognize that the corpora underlying modern LLMs are not representative of the languages used centuries ago.

The philosophy of statistical phrase alignment, albeit sparse, can shine in aligned corpora like the Rosetta Stone. Therefore, it's proposed to utilize Statistical Phrase Alignment based on token-to-token alignment rather than self-attention, along with the application of the Moses toolkit (Hoang and Kohen, 2008). Furthermore, applying the highly effective tokenization method, Byte Pair Encoding (BPE), for tokenization after alignment is suggested.

Thirdly, rather than tokenizing source and target separately, integration based on the thinking process of LLMs is proposed. Integrating source and target corpora to build dictionaries aligns with the research of Park et al. (2020). This word integration is particularly crucial in classical translation, as some expressions or terms from ancient times remain unchanged and relevant today.

## 5 Experiments

### 5.1 Statistical phrase alignment

This approach utilized the Moses toolkit implemented by Hoang and Kohen (2008). Firstly, tokenization at the character level was performed. Then, using SRILM (Stolcke, 2002), a 3-gram language model for modern Korean, the target language, was created. Next, using the training corpus, alignment at the word level between classical and Korean sentences was conducted using GIZA++ (Och and Ney, 2003), and a binary phrase table was generated.

Using the test corpus, phrase-based decoding was performed. Beam search was used to generate 1-best, n-best lists, and word lattices. These generated outputs were then compared with the ground truth data from the text corpus. The comparison was made using the BLEU module provided by Moses, resulting in a BLEU score of 0.

### 5.2 LLM in context Learning : XGLM and SOLAR

This approach utilized the open-icl framework implemented by Wu et al. (2023), which allows for comparing the performance of various models such as XGLM, LLAMA, ChatGPT, GPT-4, M2M, and NLLB using a series of templates available on Hugging Face. Among the high-performing models available for research purposes, XGLM was chosen as it is known to present superior performance in multilingual translation (Xi et al. 2022). SOLAR-10.7B, a fine-tuned version of LLAMA on Korean corpora, was also tested. As SOLAR is not incorporated in the open-icl framework, we reuse the template generated in test on XGLM. We had a total of 50,555 test corpora available, but because of lack of GPU resources and time constraints, we only tested with 1,000 data points. Testing 1,000 data points with SOLAR took around 20 hours using three RTX A5000 GPUs with each of 24GB of memory.

The newly generated sequences and target sequences was tokenized using each of XGLM and SOLAR vocabulary. We calculated BLEU score using sacReblue packages maintained by Kohen. The average standard BLEU score for sentences generated by XGLM-1.7 and SOLAR-10.7B using 20 samples with template were 1.7 and 12.7 respectively. Table 2 presents translated sentence samples.

| Type | Sample |
|---|---|
| Source | "夜流星出天厠星入南方天際狀如拳尾長四五尺許色白. |
| Target (Modern Korean) | 밤에 유성이 천측성(天厠星)에서 나와 남방 하늘가로 들어갔는데, 모양은 주먹과 같고 꼬리 길이는 4~5 척쯤 되었으며 백색이었다. |
| Target (English) | At night, a meteor came out of the astronomical star and entered the southern sky. It was shaped like a fist, had a tail about 4 to 5 feet long, and was white. |
| SOLAR prediction | 夜에 流星이 天(天厠)으로 나오고, 星(星)이 南方으로 들어오며, 天際(天 際)가 拳(拳)의 尾(尾)과 같았으나 4~5 尺(尺) |
| XGLM prediction | 무리가 졌다. 己日 = 무리가 졌다. 己日 = 무리가 졌다. 己兩司曰.. |
| MOSES-SP prediction | 밤에 유성이 天 厠 星 ) 에서 나와 남쪽 하늘가로 들어갔는데 , 모양은 주먹 같았고 꼬리의 길이는 4 ~ 5 척쯤 되었으며 백 색이었다 . |

Table 2 Samples of Translation

## 5.3 Inter Methodological Approach: Moses – SP

This approach mainly follows the classical statistical phrase alignment using Moses toolkit (Koehn et al. 2007) except for tokenizing. Before performing phrase alignment, the training and test corpora were tokenized using the BPE (Byte Pair Encoding) method in SP (sentence_piece) library, generating 10,000 tokens. These tokens were then used to tokenize both the classical and Korean languages. Moses toolkit does not utilize the GPU but only CPU. We use multi-core processing owing to m-giza framework (Och and Ney, 2003), which took less than 24 hours to train 252,773 data points with Intel(R) Xeon(R) Gold 6226R CPU s with 16 cores in total. The predictions are then scored using sacRebleu packages mentioned earlier. The average standard BLEU score is 36.71, which is higher than previous work on the very same corpus (Park et al. 2020). As can be seen in Table 2, the results of translation was satisfactory compared to other approaches.

## 6 Results Analysis

The sacRebleu package provides three kinds of scores: The standard BLEU score, calculated based on n-gram precision and brevity penalty without any smoothing. Smoothed BLEU scores which help prevent a dramatic drop in the score due to missing small n-gram, which includes Floor smoothing, Add-k smoothing (Chen et al. 2014)

The translation performance of our proposed approach Moses-SP, Solar, and XGLM is depicted in the Figure2 below.

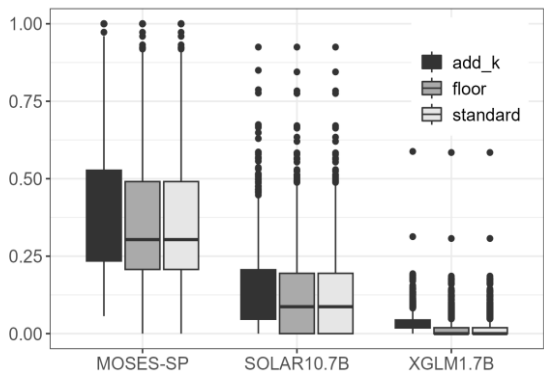

Figure 1. Comparison of model performance

The results of examining the density of standard BLEU scores by classifying ancient language characters into five categories are as follows.

The distribution of BLEU scores for sentences with character sizes ranging from 51 to 97 is closer to 0 compared to the distribution for shorter sentences. In addition, the distribution for shorter sentences is flatter in shape than the group from 51 to 97.

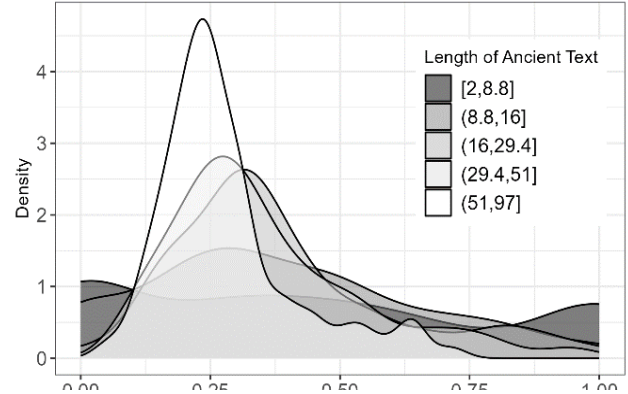

Figure 3. Distribution of Moses' prediction results of Bleu scores by length of Ancient Texts

The result patterns of SOLAR differed somewhat from the above results. Shorter sentences were translated poorly and the BLEU scores for sentences with lengths between 29.4 and 51 characters were higher compared to others. All analysis results and source codes in R and Scripts are available on author's github[2].

## 7 Conclusion

This study demonstrated how the performance of classical statistical translation techniques can be significantly improved by incorporating various recent advancements in machine translation. Although SOLAR, which was one of the comparison models in this study, showed relatively lower performance, it was evaluated using only 20 contexts, making it difficult to consider it as a direct comparison under the same conditions. Therefore, it cannot be conclusively stated that traditional statistical methods are superior to LLMs. Nonetheless, fast, and accurate statistical phrase alignment remains a crucial asset for interpreting low-resource languages such as ancient languages.

---

[2] http://github.com/wildcat842/Ancient-Korean-Archive-Translation